\documentclass{article}

\usepackage[preprint]{neurips_2026}

\usepackage[utf8]{inputenc}
\usepackage[T1]{fontenc}
\usepackage{hyperref}
\usepackage{url}
\usepackage{booktabs}
\usepackage{amsfonts}
\usepackage{nicefrac}
\usepackage{microtype}
\usepackage{xcolor}
\usepackage{graphicx}
\usepackage{amsmath}
\usepackage{amssymb}
\usepackage{placeins}
\usepackage{multirow}

\setcitestyle{numbers,square,comma}

\newcommand{\bx}{\mathbf{x}}

\newcommand{\bm}{\mathbf{m}}

\newcommand{\bs}{\mathbf{s}}

\newcommand{\bu}{\mathbf{u}}
\newcommand{\bv}{\mathbf{v}}

\newcommand{\RE}{\ensuremath{\mathbb{R}}}

\newcommand{\separen}[2]{#1 {\scriptsize$\pm$#2}}
\newcommand{\bseparen}[2]{\textbf{#1} {\scriptsize$\pm$#2}}

\title{Detection of Memorization in Diffusion Models Through Thermodynamic Singularity}

\author{%
  Iljung Kim, Kangwoo Park, Yung-Kyun Noh$\!$
  \thanks{Corresponding authors} 
  \\
  Hanyang University\\
  Seoul 04763, Rep. of Korea. \\
  \texttt{\{iljung0810,pkw3322,nohyung\}@hanyang.ac.kr} \\
  \And
  Daniel D.~Lee\textsuperscript{*} \\
  Cornell Tech \\
  New York, NY 10044, USA \\
  \texttt{ddl46@cornell.edu} \\
}

\begin{document}

\maketitle

\begin{abstract}

In diffusion models, proper distributional learning entails reversing a diffusion process over underlying density functions, whereas a key failure is memorization, in which the reverse process instead learns diffusion from individual training samples. We provide an analytical characterization of such memorized reverse processes, along with a criterion that detects the associated thermodynamic singularity. The proposed criterion reliably distinguishes between properly learned and memorizing networks across a range of neural network models designed to promote or mitigate memorization.


\end{abstract}

\section{Introduction}
\label{sec:intro}

High-fidelity generation with diffusion models has led to their widespread use, while also raising concerns about memorization. Memorization can be understood as overfitting to the forward diffusion dynamics, where the learned reverse process becomes tied to sample-specific trajectories originating from individual training samples. In the extreme, exact time-reversal of forward diffusion trajectories can reproduce training samples, resulting in limited diversity and generalization and introducing privacy risks \cite{carlini2023extracting, somepalli2023diffusion}.

Detecting such behavior is not straightforward. Direct comparison between generated samples and training data has been widely used to assess memorization \cite{somepalli2023understanding}, but it requires access to the full training dataset and a suitable similarity metric. Several proxy measures of these overfitting have also been proposed, including those based on the training loss, the Hessian spectrum of the log density \cite{jeon2025understanding}, and the difference between conditional and unconditional scores \cite{wen2024detecting}. However, the measures based on these proxies are largely empirically motivated and rely on high-dimensional estimation, making them difficult to interpret in a principled manner and sensitive to noise.

In principle, these trajectories should reveal clear and observable differences between overfitted and properly learned networks. However, such differences diminish as the number of training samples grows, since the trajectory structure induced by individual samples becomes progressively smeared. Nonetheless, retrieval of training samples and the generation of new samples remain fundamentally different, and detecting the difference is important even in large datasets, as the possibility of reproducing training samples---even if small---can severely limit the applicability of the model.

In this paper, we provide a simple derivation using the equation of continuity, showing that the velocity field diverges in the diffusion of Dirac delta functions. This singularity, which can serve as a necessary condition for memorization, arises in all forward processes constructed from finite samples, regardless of sample size. Memorization in the reverse process cannot occur without learning this singularity, and this singularity provides a direct analytical characterization of the presence of memorization.

The mathematical analogy of the velocity field and its corresponding flux naturally leads to a thermodynamic quantity---the Helmholtz free energy of the diffusive process. The memorization characteristic can be interpreted using the instantaneous rate of change of the Helmholtz free energy over time, providing a principled and interpretable scalar measure that does not require access to training data. Using this rate of change as a criterion, the detection of memorization was effective across both overfitting and non-overfitting neural networks, outperforming conventional methods on synthetic and real-world datasets. In addition, we further propose a simple modification of the training loss function that suppresses excessive growth of the diffusive velocity above a predefined threshold based on the free energy rate, significantly reducing memorization without degrading the fidelity of generated samples.

The remainder of the paper is organized as follows. Section 2 derives the low-noise behavior of velocity fields associated with diffused delta functions. Section 3 introduces the thermodynamic criterion and its interpretation. Section 4 presents experiments on synthetic and real-world datasets, including memorization mitigation. Section 5 concludes.

\section{Forward and Reverse Processes in Diffusion Models}

The forward process in diffusion models adds noise to all data using the following update equation:
\begin{eqnarray}
    \mathbf{Forward\ process\!:}\quad \bx_{t+\Delta t} = \sqrt{1 - \beta\Delta t}\ \bx_t + \sqrt{\beta\Delta t}\ \epsilon_t, \quad \epsilon_t\sim\mathcal{N} (\epsilon_t; 0, I), \label{eq:forward_pr}
\end{eqnarray}
for $\bx_t, \epsilon_t\in\RE^d$.
The reverse process is augmented by an additional term $\bu_t(\bx_t)$ for control:
\begin{eqnarray}
    \mathbf{Reverse\ process\!:}\quad \bx_{t-\Delta t} = \bx_t  + \bu_t(\bx_t)\Delta t + \sqrt{\beta\Delta t}\ \epsilon_t, \quad \epsilon_t\sim\mathcal{N}(\epsilon_t; 0, I).
\end{eqnarray}
The control term $\bu_t(\bx_t)$ incorporates neural networks in various forms, which are designed to reverse the diffusive flow of the forward process using the information learned during that process.

\textbf{Score field:} The diffusive flow induced by the diffusion of a density $\rho_t(\bx)$ can be explained using the convolution with a small Gaussian kernel,  $k_{\Delta t}(\bx) = \mathcal{N}(\bx; 0, \beta\Delta tI)$. The density after a small diffusion step $\Delta t$ can be written as
$\rho_{t+\Delta t}(\bx) = \int k_{\Delta t}(\bx, \vec{0})\rho_t(\bx)d\bx' 
\approx \rho_t(\bx) + \frac{\beta\Delta t}{2}\nabla^2\rho_t$, yielding
\begin{eqnarray}
    \frac{d\rho_t}{dt} = \lim_{\Delta t\rightarrow0}\frac{\rho_{t+\Delta t}(\bx) - \rho_t(\bx)}{\Delta t} \approx \frac{\beta}{2}\nabla^2\rho_t.
    \label{eq:rho_rate}
\end{eqnarray}
Fick's law, $J = -D\nabla\rho_t$, describes the diffusive flux $J (=\rho_t\bv)$ driven by the density gradient, with a constant coefficient $D$. Along with the equation of continuity $\frac{\partial\rho_t}{\partial t} = -\nabla\cdot J$, we can produce the equation $\frac{\partial \rho_t}{\partial t} = -\nabla\cdot J = D\nabla^2\rho_t$ which corresponds to Eq.~(\ref{eq:rho_rate}). The correspondence yields the constant $D = \frac{\beta}{2}$. Then, from the definition of flux, $J=\rho_t\bv$, we can obtain the diffusive velocity field:
\begin{eqnarray}
    \bv(\bx) = \frac{J}{\rho_t} = -D\frac{\nabla\rho_t}{\rho_t} = -\frac{\beta}{2}\nabla\log\rho_t,
    \label{eq:diffusion_velocity}
\end{eqnarray}
where $\nabla\log\rho_t$ is the score vector of a density $\rho_t(\bx)$. The score field in a pure diffusion system is proportional to the velocity field induced by diffusion.

For the forward process, along with that $\lim_{\Delta t\rightarrow0}\frac{\bx_{t+\Delta t} - \bx_t}{\Delta t} = \lim_{\Delta t \rightarrow 0} \frac{\sqrt{1 - \beta\Delta t}\ \bx_t - \bx_t}{\Delta t} + o(\Delta t) = -\beta/2\bx_t$, it induces a velocity field in the infinitesimal limit $\Delta t\rightarrow0$:
\begin{eqnarray}
    \bv(\bx) = -\frac{\beta}{2}\left(\bx + \nabla\log\rho_t \right), \label{eq:velocity_forward}
\end{eqnarray}
where the two terms arise from the corresponding components in the update rule in Eq.~(\ref{eq:forward_pr}). In general, neural networks are designed to directly predict a score $\nabla\log\rho_t$ or the expected total accumulated noise $\epsilon_{\mathrm{tot}, t}$, which corresponds to $\epsilon_{\mathrm{tot}, t} = -\sigma_t\nabla\log\rho_t$, with $\sigma_t=\sqrt{1-e^{-\beta t}}$ or $\sigma_t = \sqrt{1 - (1 - \beta\Delta t)^{t/\Delta t}}$ depending on continuous- or discrete-time formulation.

When the neural network overfits to approximate the exact reverse process, it learns the control field $\bu_t(\bx) = \frac{\beta}{2}\bx + \beta\nabla\log\rho_t$ which induces the reverse-process velocity field: $\bu_t -\frac{\beta}{2}\nabla\log\rho_t = -\bv(\bx)$. If the reverse process is performed without noise, i.e., $\bx_{t-\Delta t} = \bx_t + \bu_t(\bx_t)\Delta t$ without noise addition, the control field is simply $\bu_t(\bx) = \frac{\beta}{2}(\bx + \nabla\log\rho_t) = -\bv_t$.

\subsection{Diffusion of $\delta$-Functions}
\label{sec:delta_function_diffusion}

A surprising yet general property of the forward process emerges from deriving the induced velocity field associated with the diffusion of a Dirac $\delta$-function. Consider a Gaussian kernel $k_{t}(\bx, \bx') = \mathcal{N}(\bx; \bx', t\beta I)$ which satisfies the convolution property: ${\color{black}
k_{t + \Delta t}(\bx,\bx') = \int k_{\Delta t}(\bx')k_t(\bx,\bx') d\bx'
}$.

The density recovers the Dirac $\delta$-function in the limit $\lim_{t\rightarrow0}k_{t}(\bx, \bx') = \delta(\bx-\bx')$. The induced velocity field $\bv_t(\bx)$ associated with the diffusion satisfies the continuity equation:
\begin{eqnarray}
    \nabla\cdot\Big( \bv(\bx) k_{t}(\bx, \bx')\Big) = -\frac{\partial k_{t}(\bx, \bx')}{\partial t}.
\end{eqnarray}
with the flux $J(\bx)=\bv(\bx) k_t(\bx, \bx')$. Expanding both sides yields
\begin{eqnarray}
    \left\{ \nabla\cdot\bv - \frac{\bx-\bx'}{t\beta}\cdot\bv(\bx) \right\}k_{t}(\bx, \bx') = -\frac{1}{2t}\left( \frac{(\bx-\bx')^2}{t\beta} - d\right)k_{t}(\bx, \bx'),
\end{eqnarray}
The plug in of an equation of the form $\bv(\bx)\equiv(\bx-\bx')f(t)$ for some scalar function $f(t)$, we obtain $f(t) = \frac{1}{2t}$, which yields the velocity field associated with this vanilla diffusion:
\begin{eqnarray}
\boxed{
\bv(\bx) = \frac{1}{2t}(\bx - \bx').
}
\end{eqnarray}
Note that the velocity field "diverges" as $t\rightarrow0$ everywhere, indicating that the diffusion of a Dirac $\delta$-function induces an infinite initial velocity, which subsequently decays proportionally to $1/t$. Note that this pure diffusive property of the velocity does not depend even on the noise magnitude $\beta$.


Extensions of this single $\delta$-function derivation to mixtures of $\delta$-functions are straightforward. In this view, training samples can be interpreted as a mixture of Dirac delta functions. Since the velocity grows linearly with distance from each center, while Gaussian densities decay exponentially, the resulting flux is dominated by the Gaussian mixture structure. Consequently, the velocity field at any point is largely determined by the nearest center. As long as the diffused Gaussians are not sufficiently smeared, the singular behavior of the initial velocity field persists throughout the space. In diffusion models, when centers correspond to training samples, the velocity field diverges in the early stage of the forward process; moreover, due to $\bv \propto \bx - \bx'$, this divergence is more pronounced farther from the centers than nearby, which may appear counterintuitive.

Our key inference for detecting memorization is as follows. Networks trained on finite samples in the forward process induce singularities in the early stage. The absence of such singularities in the final stage of the reverse process indicates that the network has not memorized.

\section{Criterion for Detecting Memorization}
\label{sec:related_work}

Memorization has been identified by examining nearest neighbors of generated samples within the training dataset. Detecting duplication of training data via nearest-neighbor comparisons is considered a gold--standard indicator of memorization \cite{carlini2023extracting, somepalli2023diffusion, somepalli2023understanding}. However, this approach requires access to the full training dataset, which limits its practical applicability.

Instead, memorization can be investigated through intrinsic properties of the learned neural networks. Because gold-standard indicators can, even in large datasets, unambiguously distinguish memorizing networks from non-memorizing ones, it is natural to seek corresponding signatures within trained networks. Existing approaches pursue this direction by introducing empirical proxies for memorization.

These approaches include guidance-based methods that identify the explosion of guidance (showing large magnitude in  the difference between conditional and non-conditional scores) under the premise that such behavior sharpens the conditioned probability landscape \cite{wen2024detecting}, attention-based methods that identify anomalous late-stage attention behavior \cite{chen2025bright}, and geometrical approaches that detect sharp and highly localized regions encoded in the Hessian of the log density \cite{jeon2025understanding}. Notably, these methods, to varying degrees, attempt to analytically relate their criteria to the norm of the score function, motivated by empirical observations of irregular score behavior in memorizing networks, which remain only partially understood. These works interpret the empirical growth of score magnitudes as a consequence of the criteria they propose as principled.

On the contrary, we argue that the explosion of the score itself constitutes a central identifier of memorization. Specifically, within the diffusion model pipeline, memorization consistently occurs in conjunction with the explosion of the velocity over the entire space.  Consequently, in the absence of large velocity values, memorization cannot occur. Therefore, the absence of large velocity values intuitively indicates non-memorization, while the converse does not necessarily hold. Furthermore, suppressing large velocity magnitudes---either explicitly or implicitly, e.g., via architectural constraints of neural networks or early stopping—will lead to the inhibition of memorization.


\subsection{Helmholtz Free Energy Rate Criterion}

We propose a criterion for assessing the non-memorization behavior of neural networks in diffusion models:
\begin{eqnarray}
\boxed{
\Phi \equiv \frac{1}{T}\!\left.\widehat{\frac{d F}{d t}}\right|_t = -\frac{\beta}{2} \frac{1}{N}\sum_{i = 1}^N ||\widehat{\bv}_t(\bx_{t,i})||^2.
} \label{eq:proposed_criterion}
\end{eqnarray}
The samples $\bx_{t,i}$ are the $i$-th sample at time $t$ obtained during the reverse process, and the vector $\widehat{\bv}_t(\bx)$ is defined based on the output of the learned neural network. Specifically, we use its form $\widehat{\bv}_t(\bx) = \bx - \widehat{\epsilon}_{\mathrm{tot}, t}^\theta(\bx)/\sigma_t$ when the network $\widehat{\epsilon}_{\mathrm{tot}, t}^\theta(\bx)$ learns accumulated noise, and $\widehat{\bv}_t(\bx) = \bx + \widehat{\bs}_t^\theta(\bx)$ when $\widehat{\bs}_t^\theta(\bx)$ learns the score. $F$ is the Helmholtz free energy which will be introduced later.

\begin{figure}[t]
    \centering
    \includegraphics[width=0.94\textwidth]{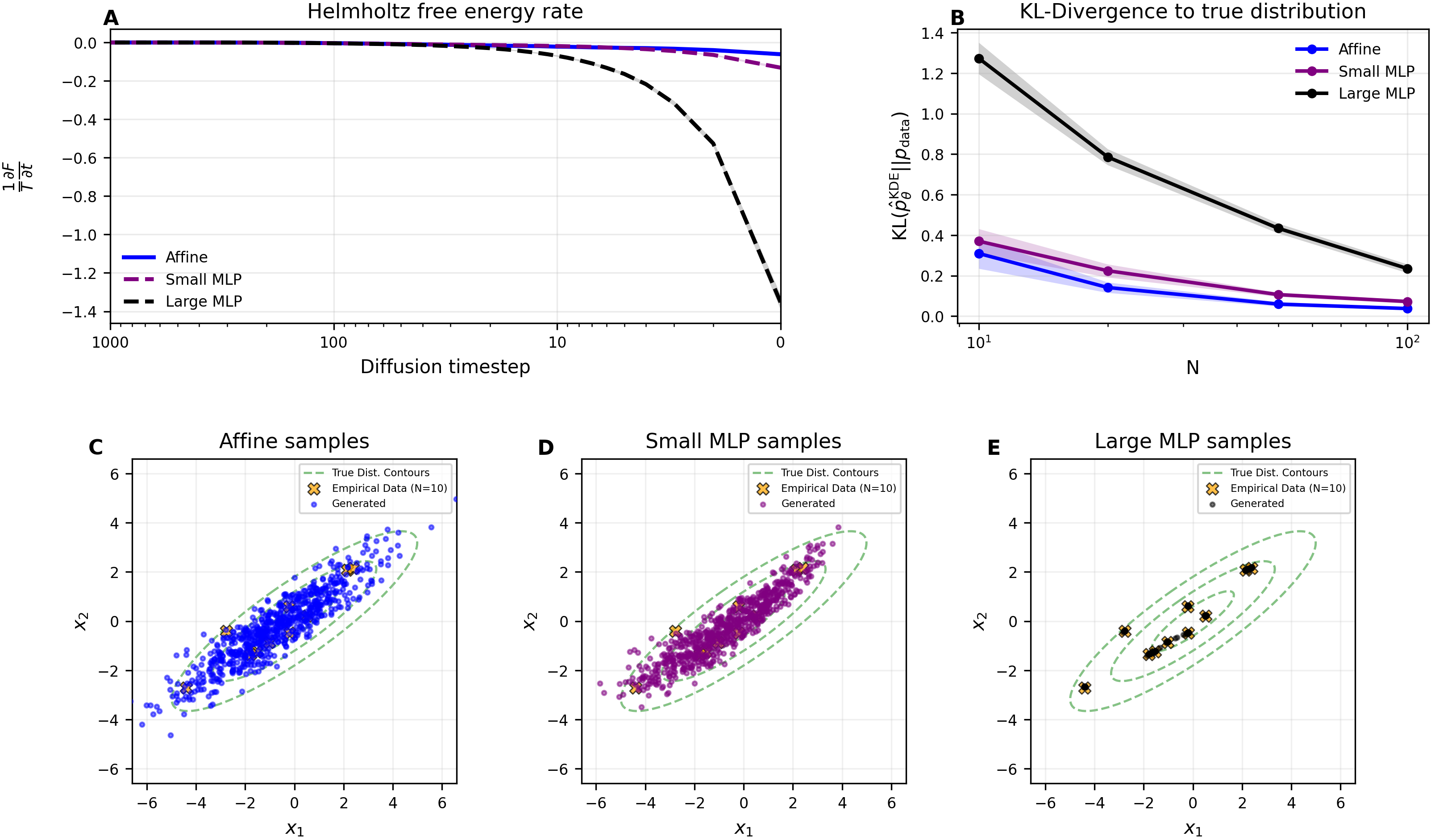}
    \vspace{-0.75em}
\caption{
\textbf{2D Gaussian generation.}
(A) The rate of Helmholtz free energy dissipation for an affine network, a small MLP, and a large MLP, plotted on a logarithmic scale over time steps.
(B) K-L divergence between the generated samples and the true density as a function of training sample size.
(C--E) Generated samples from the affine network, small MLP, and large MLP models, respectively. After sufficient training, the failure of the low-capacity model to capture free energy dissipation until a certain time leads to the successful reproduction of the underlying Gaussian structure.
}
\label{fig:gaussian_experiment}
\end{figure}

\subsection{Behavior of the Criterion for Two-Dimensional Gaussian Generation}
\label{subsec:gaussian_experiment}

Previously, various explanations have been proposed for why diffusion models tend to generalize, including early stopping \cite{bonnaire2025diffusion,NEURIPS2023_06abed94}, network architecture \cite{kadkhodaie2024generalization}, and optimization methods such as stochastic gradient descent and learning rate \cite{wu2025bigstep}. We consider an affine neural network. In our example, the network takes the form
\begin{eqnarray}
\bs_\theta(\bx, t)=A_\theta(t)(\bx - \mu_\theta(t)),    
\end{eqnarray}
where $A_\theta(t)\in\RE^{d\times d}$ is a symmetric matrix and $\mu_\theta(t)\in\RE^d$ is a vector. The $(d+1)d/2$ independent entries of $A_\theta(t)$ and the $d$ entries of $\mu_\theta(t)$ are parameterized as outputs of neural networks. A simple derivation shows that this network recovers the score corresponding to a Gaussian with maximum-likelihood estimates of the mean and covariance. The full derivation is provided in Appendix~\ref{app:gaussian_details}. 

We hypothesize that the large multilayer perceptron (MLP) overfits and reproduces the training samples, the small MLP generates samples near the training data, and the aforementioned affine network produces Gaussian samples with the empirical mean and covariance. Using the forward and reverse processes, we analyze movement of criterion, or the dissipation rate of the Helmholtz free energy, induced by each model. For our synthetic experiment, we generated a small dataset ($N = 10$) from a Gaussian and trained diffusion models using an MLP with both large and small capacities, as well as an affine model. 

We could expect that the forward diffusive field initialized from the 10 samples exhibits  singular behavior in the initial velocities. As shown in Figure~\ref{fig:gaussian_experiment}(A), the large MLP appears to have learned detailed trajectories, moving from $t=1{,}000$ to $t=0$ during the reverse process, including the singular behavior ($\Phi=-\infty$) of the initial diffusion around $t=0$.

In contrast, the small MLP and the affine model do not follow this dissipation behavior; their dissipation rates stop decreasing around $t=200$. The affine model, being well-aligned with the true underlying density, produces samples that closely match the true density, as shown in the K-L divergence estimation in Figure~\ref{fig:gaussian_experiment}(B). The generated samples in Figure~\ref{fig:gaussian_experiment}(C--E) illustrate that the affine model produces Gaussian samples with the empirical mean and covariance of the 10 training points. The small MLP exhibits partial generalization beyond the training data, whereas the large MLP memorizes the training set, accompanied by a pronounced decay in the Helmholtz free energy dissipation rate.

In Figure~\ref{fig:gaussian_experiment}(A), we also observe that Helmholtz free energy dissipation rate is negative and monotonically decreases throughout the reverse process. Including these properties, we next elucidate how the norm of the score is intimately connected to a thermodynamic quantity, shedding light on the mechanism that links large score magnitudes to a principled theoretical criterion.

\subsection{Helmholtz Free Energy in Diffusion Models}

The forward and reverse processes in diffusion can be interpreted within a thermodynamic framework, where their states are represented by the following thermodynamic functional:
\begin{eqnarray}
F[\rho_t] = \langle U\rangle_{\rho_t} - TS = \! \int\! \rho_t(\bx) U(\bx) d\mathbf{x} + T\int \!\rho_t(\bx) \log \rho_t(\bx) d\mathbf{x}.
\end{eqnarray}
Here, $\rho_t(\bx)$ denotes the probability density that changes over time, $U(\bx)$ represents the energy function, and $T$ denotes the fixed temperature in the system. The quantity $\langle U\rangle_{\rho_t}$ denotes the expectation of $U(\bx)$ with respect to $\rho_t$, and $F[\rho_t]$ is the Helmholtz free energy with the Boltzmann constant is set to unity.

Given $U(\bx)$, the thermodynamic law decreases $F[\rho_t]$ over time by updating $\rho_t(\bx)$ according to the equation of continuity. The rate of change of the Helmholtz free energy is represented using the change with respect to $\rho_t$ by the chain rule:
\begin{eqnarray}
    \frac{d F}{d t} = \left\langle \frac{\partial F}{\partial \rho_t}, \frac{\partial \rho_t}{\partial t} \right\rangle 
    = \int \frac{\partial F}{\partial \rho_t} \, \frac{\partial \rho_t}{\partial t} \, d\mathbf{x},
\end{eqnarray}
where $\left\langle f_1, f_2 \right\rangle$ denotes the functional inner product of two functions $f_1(\bx)$ and $f_2(\bx)$, and both $\frac{\partial F}{\partial \rho_t}$ and $\frac{\partial \rho_t}{\partial t}$ are pointwise functions of $\bx$.
Applying the equation of continuity and integration by parts, we obtain
\begin{eqnarray}
    \int \left.\frac{\partial F}{\partial \rho_t}\right|_{\rho_t}\frac{\partial \rho_t}{\partial t} d\bx &=& -\int \left.\frac{\partial F}{\partial \rho_t}\right|_{\rho_t}\nabla\cdot(\rho_t\bv) d\bx \\
    &=& \int \rho_t \bv(\bx)^\top \nabla\!\left.\frac{\partial F}{\partial \rho_t}\right|_{\rho_t} d\bx \\
    &=& \left\langle \bv, \nabla\frac{\delta F}{\delta \rho_t} \right\rangle_{\rho_t}, \label{eq:dFdt}
\end{eqnarray}
where $\langle\bv_1,\bv_2\rangle_{\rho_t}$ denotes the expectation of $\bv_1(\bx)^\top\bv_2(\bx)$ under $\rho_t$. Given a fixed kinetic energy $K=\frac{1}{2}\int\rho_t||\bv||^2d\bx$, the velocity field that maximizes the reduction of Helmholtz free energy  is $\bv \propto -\nabla\!\left.\frac{\delta F}{\delta \rho_t}\right|_{\rho_t}$, and the magnitude of the dissipation rate $\frac{d F}{d t}$ becomes proportional to the kinetic energy. The theoretical significance of the free energy lies in the fact that its maximal rate explains the thermodynamic motion observed in natur de, analogous to the principle of least action.

Because $\frac{\partial F}{\partial \rho_t} = U + T(1 + \log\rho_t)$ at each $\bx$, the velocity and the flux of the system are
\begin{eqnarray}
    \bv(\bx) &=& -\nabla\frac{\delta F}{\delta \rho_t} \ =\  -\nabla U -T\nabla\log\rho_t \\
    J(\bx) &=& \rho_t(\bx)\bv(\bx) \ =  \ -\rho_t(\bx)\nabla U - T\nabla\rho_t.
\end{eqnarray}
For the correspondence to the velocity field of diffusion models in Eq.~(\ref{eq:velocity_forward}), we use the temperature $T=\frac{\beta}{2}$, and the energy for control is $U(\bx) = \frac{\beta}{4}\bx^2$ from $\nabla U = \frac{\beta}{2}\bx$. 

The forward process behaves under a fixed control potential $U(\bx)=\frac{\beta}{4}\bx^2$. At equilibrium, the probability flux vanishes, i.e., $J = -\rho_\mathrm{eq}\nabla U - T\nabla\rho_\mathrm{eq} = 0$. We can write the equation $\nabla\rho_\mathrm{eq} = -\frac{\nabla U}{T}\rho_\mathrm{eq}$, which yields the equilibrium density after sufficient forward updates:
\begin{eqnarray}
    \rho_\mathrm{eq}(\bx) \propto \exp(-U(\bx)/T) = \exp\left(-\bx^2/2\right).
\end{eqnarray}
The forward process therefore simply corresponds to the evolution of $\rho_t(\bx)$ from a mixture of Dirac $\delta$-functions to an isotropic single Gaussian with unit variance. The trajectories and velocities jointly maximize a single quantity, the rate of decrease of the Helmholtz free energy under the equation of continuity. The following summarizes the behavior of Helmholtz free energy rate during the forward and reverse processes.

\textbf{Helmholtz free energy does not increase:} With the velocity field $\bv(\bx) = -\nabla\frac{\delta F}{\delta \rho_t} = -\nabla U - T\nabla\log\rho_t$, the Helmholtz free energy is non-increasing in time, since $\frac{d F}{d t} = -\langle\bv, \bv\rangle_{\rho_t} \leq 0$ from Eq.~(\ref{eq:dFdt}). When $\rho_t(\bx)$ reaches the equilibrium density, the flux vanishes, and the free energy no longer decreases. Throughout the forward process, the free energy continuously decreases under a fixed potential $U(\bx) = \frac{\beta}{4}\bx^2$. During the reverse process, the potential becomes time-dependent to induce the control $\bu_t(\bx)$. Nevertheless, the rate of free energy dissipation remains unchanged for a fixed $\bu_t(\bx)$ at time $t$, since the velocity field is identical up to a sign reversal.

\textbf{The rate of change decreases over time during the forward process:} The second derivative $\frac{\partial^2 F}{dt^2}$ is positive, as implied by results on the dissipation of Fisher information in \cite{Villani2009,BakryEmery1985} provided that the potential is convex. Under the convex potential $U(\bx) = \frac{\beta}{4}\bx^2$, the initially large rate of change does not persist over time during the forward process. The large rate of change occurs only when $t$ is small.

From these two theoretical properties, the proposed criterion is always negative, and its magnitude does not increase during the forward process. The absence of these properties in the criterion during the reverse process indicates that the network has not overfit the forward diffusive flow and can be used as a criterion for ensuring non-memorization.

In the experiments, we evaluate the effectiveness of the proposed criterion in detecting memorization in neural networks, as well as memorization at the level of individual samples. We also incorporate the criterion into the objective function to encourage the model to generate outputs without memorization.

\section{Experiments}
\label{sec:experiments}

In this section, we evaluate the criterion proposed in Eq.~(\ref{eq:proposed_criterion}) across various benchmark datasets and discuss the results. While the criterion is computed at every step of the reverse data-generating process using trained neural networks, in practical applications it can be evaluated at a single time step---specifically at the final stage of generation. We use pretrained networks for CIFAR-10 and CelebA-HQ, and then induce catastrophic corruption by  fine-tuning them on a small number of samples. We also evaluate the individual score to assess the degree of memorization of each generated sample. Finally, we incorporate the proposed criterion into the training loss and examine whether it mitigates memorization while preserving the quality of generated samples. Details of the experiment are explained in Appendix~\ref{app:experiment_details}.

\begin{figure}[t]
    \centering
    \includegraphics[width=0.94\textwidth]{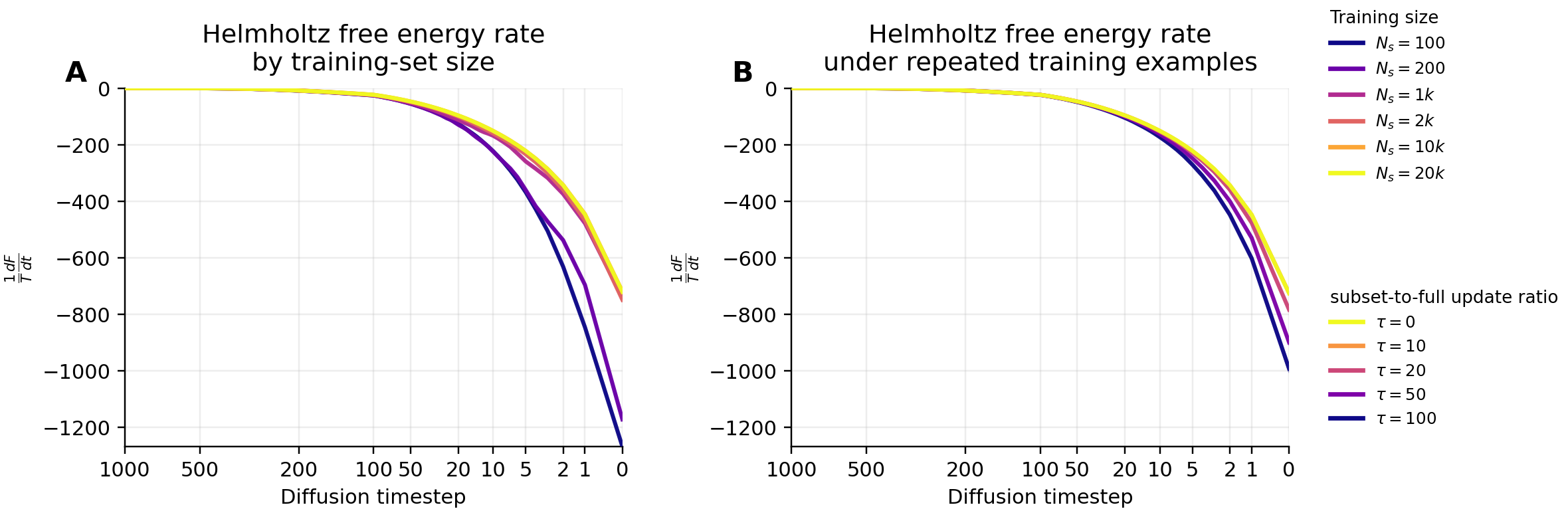}
    \vspace{-1.0em}
    \caption{
    \textbf{CIFAR-10 pixel-space fine-tuning.}
    (A) Catastrophic corruption. Pretrained networks are fine-tuned using a fixed subset of size $N_s$.
    (B) Non-catastrophic corruption. Pretrained networks are fine-tuned using a fixed subset of size $N_s=100$, with occasional training on additional samples from the full dataset.
    }
    \label{fig:cifar_finetuning}
\end{figure}

\begin{figure}[t]
    \centering
    \includegraphics[width=\textwidth]{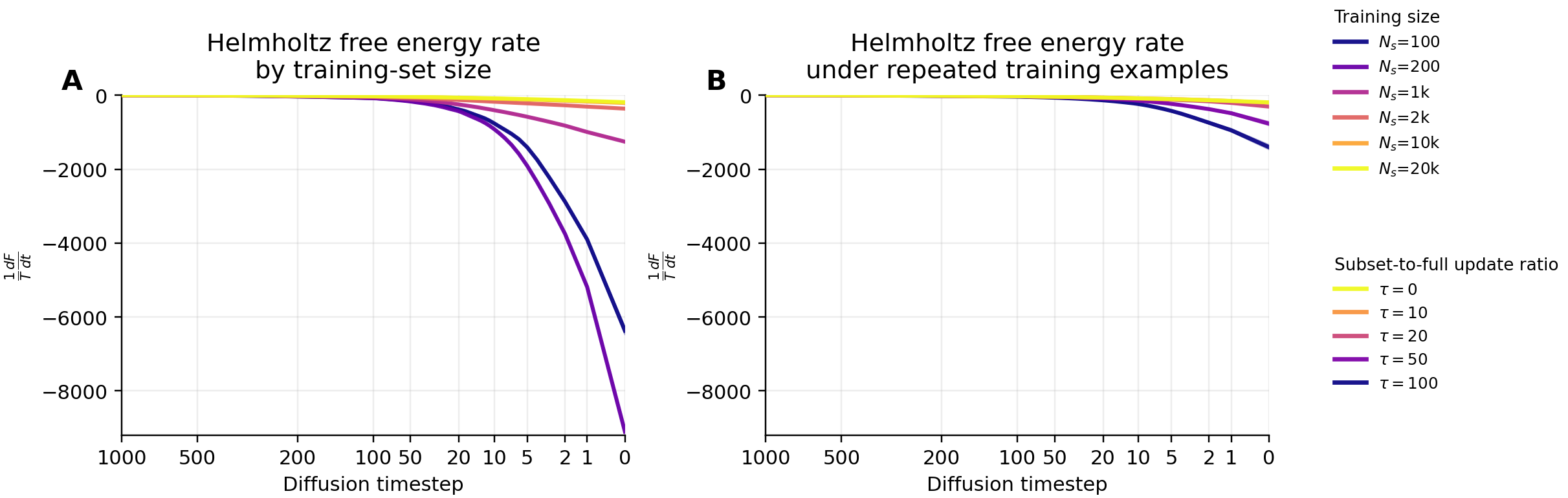}
    \vspace{-2.0em}
    \caption{
    \textbf{CelebA-HQ latent-space fine-tuning.}
    (A) Catastrophic corruption.
    (B) Non-catastrophic corruption  with occasional training on additional samples from the full dataset.
    }
    \label{fig:celeba_ldm}
\end{figure}

\subsection{Catastrophic and Non-Catastrophic Corruptions}
\label{subsec:image_diffusion}

\vspace{-0.55em}

Pretrained networks for real-image generation were used: a CIFAR-10 pixel-space diffusion model \cite{krizhevsky2009learning} and a CelebA-HQ latent diffusion model \cite{liu2015deep,karras2018progressive,rombach2022high}. These models were fine-tuned using subsets of the training data. For the catastrophic corruption experiment, we selected subsets of varying sizes $N_s$ and performed 50,000 fine-tuning epochs using only those subsets. For the non-catastrophic corruption setting, we fixed the subset of sample size $N_s=100$. After $\tau$ steps of training on this subset, we randomly sampled 5,000 additional data from the full dataset and performed one epoch of fine-tuning using these samples. We evaluated the criterion $\Phi$ throughout the reverse process in Figure~\ref{fig:cifar_finetuning} and \ref{fig:celeba_ldm}, where darker curves correspond to more memorizable setting. The CIFAR-10 model operates in pixel space, whereas the CelebA-HQ model operates in latent space. Note that the time scale in all graphs is logarithmic.

\begin{figure}[t]
    \centering
    \makebox[\textwidth][c]{%
        \includegraphics[width=\textwidth]{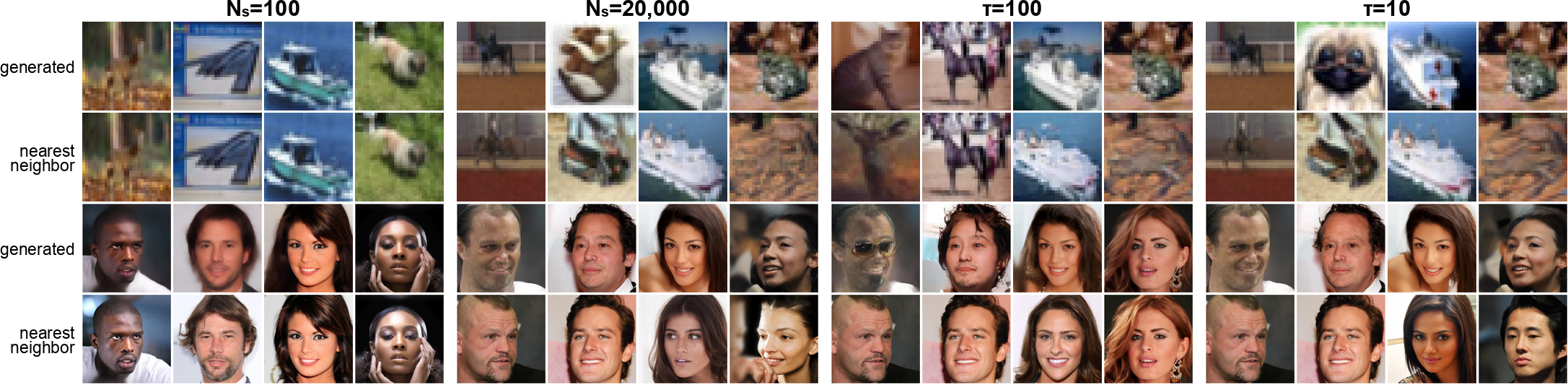}
    }
    \vspace{-1.75em}
    \caption{
    \textbf{Generated samples from fine-tuned networks and their nearest neighbors.} The first two rows show CIFAR-10 generated samples and their nearest neighbors in the training data, while the next two rows show CelebA-HQ generated samples and their nearest neighbors. We present samples obtained after catastrophic corruption with $N_s=100$ and $N_s=20{,}000$, as well as after non-catastrophic corruption with subset training for $\tau=100$ and $\tau=10$ consecutive epochs. Ordered from least to most memorization, the settings are $N_s = 20{,}000$, $\tau = 10$, $\tau = 100$, and $N_s = 100$.
    }
    \label{fig:nearest_neighbor_examples}
    \vspace{-0.55em}
\end{figure}

In these experiments, nearest neighbor LPIPS scores for the generated samples are reported in Table~\ref{tab:lpips_memorization}. 
Entries indicate the percentage of $1{,}024$ generated samples whose nearest training image falls below a specified LPIPS threshold. CIFAR-10 uses LPIPS $\le 0.03$ to avoid false positives due to low resolution, while CelebA-HQ uses LPIPS $\le 0.1$. Smaller $N_s$ and larger $\tau$ lead to increased memorization during fine-tuning. Representative generated examples and their nearest neighbors are shown in Figure~\ref{fig:nearest_neighbor_examples}. In the catastrophic corruption setting, all generated samples are memorized when $N_s=100$, whereas no samples exhibit memorization when $N_s=20{,}000$. In non-catastrophic setting, both memorized and non-memorized samples appear; however, the parameter $\tau$ explains the proportion of memorization. These observations are consistent with the criterion at $t=0$ across all cases in Figure~\ref{fig:cifar_finetuning} and \ref{fig:celeba_ldm}. These observations regarding the extent of memorization are consistent with the behavior of the criterion near $t=0$ in these figures.

\begin{table}[t]
    \centering
    \footnotesize
    \setlength{\tabcolsep}{5pt}
    \renewcommand{\arraystretch}{1.08}
    \caption{
    \textbf{Memorization with fine tuning.}
    Percentage of memorized samples during fine-tuning whose nearest neighbor in the training set falls below the LPIPS threshold. $N_s$ denotes the subset size used in catastrophic corruption, and $\tau$ denotes the number of consecutive epochs for which the $N_s=100$ subset is used during fine-tuning.
    }
    \vspace{-0.45em}
    \label{tab:lpips_memorization}
    \begin{tabular}{rcc@{\hspace{1.4em}}rcc}
        \toprule
        \multicolumn{3}{c}{\textbf{Catastrophic Corruption}} &
        \multicolumn{3}{c}{\textbf{Non-Catastrophic Corruption}} \\
        \cmidrule(lr){1-3}\cmidrule(lr){4-6}
        $N_s$ & CIFAR-10 & CelebA-HQ &
        $\tau$ & CIFAR-10 & CelebA-HQ \\
        \midrule
        100 & 99.80\% & 57.71\% & 0 & 0.00\% & 0.00\% \\
        200 & 94.53\% & 8.01\% & 10 & 0.78\% & 0.00\% \\
        1,000 & 3.71\% & 0.00\% & 20 & 5.08\% & 0.00\% \\
        2,000 & 0.78\% & 0.00\% & 50 & 18.95\% & 6.35\% \\
        10,000 & 0.49\% & 0.00\% & 100 & 32.52\% & 25.88\% \\
        20,000 & 0.39\% & 0.00\% & -- & -- & -- \\
        \bottomrule
    \end{tabular}
\end{table}

\vspace{-0.55em}

\subsection{Sample-Level Memorization}
\label{subsec:sample_level_detection}

\vspace{-0.5em}

We evaluated the score norm obtained as a byproduct of sample generation. The criterion was calculated using the conditional and unconditional velocities, $\widehat{\bv}_\mathrm{cond}(\bx)$ and $\widehat{\bv}_\mathrm{uncond}(\bx)$, respectively, at the end of the reverse-process trajectory. This sample-level metric quantifies the magnitude of the prompt-conditioned score via $||\widehat{\bv}_\mathrm{cfg}(\bx_t, t)||^2 - ||\widehat{\bv}_\mathrm{uncond}(\bx_t, t)||^2$ where $\widehat{\bv}_\mathrm{cfg}(\bx_t, t) = \widehat{\bv}_\mathrm{uncond}(\bx_t, t) + w(\widehat{\bv}_\mathrm{cond}(\bx_t, t) - \widehat{\bv}_\mathrm{uncond}(\bx_t, t))$, and $w$ is classifier-free guidance scale. Given a score $\bs_\theta(\bx_t, t)$, $\widehat{\bv}(\bx_t,t)=x_t+\bs_\theta(\bx_t,t)$.

We compare the sample-level detection performance of this criterion with the score-difference norm~\cite{wen2024detecting} and Hessian-sharpness baseline~\cite{jeon2025understanding}.
Memorized samples are defined as samples generated by controlled prompts for the purpose of memorization. The results in Table~\ref{tab:sample_level_auroc_sd14} show that the norm of the individual score serves as a reliable indicator for identifying safe generation without memorization. The experiments were conducted using Stable Diffusion v1.4 \cite{rombach2022high}.

\begin{table}[t]
\centering
\footnotesize
\setlength{\tabcolsep}{5pt}
\renewcommand{\arraystretch}{1.12}
\caption{ \textbf{Sample-level detection on SD v1.4.}
The generated samples from memorizing prompts and non-memorizing prompts are discriminated by various memorization criteria. 
All measurements are performed on DDPM 1,000-step trajectories.
Bold entries are statistically indistinguishable from the best value in each column under an approximate two-sided test. Time step 1,000 corresponds to the beginning of diffusion, and time step 1 to its end.
}
\vspace{-0.45em}
\label{tab:sample_level_auroc_sd14}
\begin{tabular}{lccc}
\toprule
\textbf{Method}
& \textbf{Time step}
& \textbf{AUC}
& \textbf{TPR@1\%FPR} \\
\midrule

$\left\|\bs_\theta^\Delta(\bx_t)\right\|$ 
& 1,000
& \separen{0.937}{0.012}
& \separen{0.771}{0.027} \\

$ = \|\bs_{\theta,\mathrm{cond}}-\bs_{\theta,\mathrm{uncond}}\|$ \cite{wen2024detecting}
& 1
& \separen{0.972}{0.005}
& \bseparen{0.909}{0.020} \\

\midrule

$\left\|H_\theta^\Delta(\bx_T)\bs_\theta^\Delta(\bx_T)\right\|^2$ \cite{jeon2025understanding}
& 1,000
& \separen{0.962}{0.008}
& \separen{0.856}{0.024} \\

& 1
& \separen{0.970}{0.005}
& \bseparen{0.874}{0.021} \\

\midrule

$\|\widehat{\bv}_{\mathrm{cfg}}\|^2-\|\widehat{\bv}_{\mathrm{uncond}}\|^2$ (Ours)
& 1
& \bseparen{0.982}{0.003}
& \bseparen{0.905}{0.019} \\

\bottomrule
\end{tabular}
\end{table}

\vspace{-0.5em}

\subsection{Loss Function Modification}
\label{sec:mitigating_memorization_reference_scale}

\vspace{-0.5em}


We examined the incorporation of the criterion into the loss function to assess its ability to improve fidelity without inducing memorization. With a threshold 
$R_t=(\beta_t/2)\,\mathbb{E}_{\mathrm{point}}[\|\widehat{\bv}_t\|_2^2] \simeq \frac{\beta_t(1-\sigma_t^2)}{2\sigma_t^2}d$, we penalize the norm along the trajectory for points whose values exceed $\gamma R_t$, where $0<\gamma\leq1$ is a control parameter. This discourages the neural network from learning excessively large scores. Here, $d$ denotes the dimensionality, and $\sigma_t^2$ is the accumulated variance defined as $\sigma_t^2 = \left(1 - \prod_{s=0}^{t-1}(1 - \beta_t)\right)$, where $\beta_t$ denotes the noise schedule $\beta$ at each time step $t$.

In the CelebA-HQ non-catastrophic corruption setting described in Section~\ref{subsec:image_diffusion}, the latent diffusion model is fine-tuned for $50{,}000$ steps starting from a pretrained model with a subset-to-full update ratio of $\tau=100$. For samples whose score norm exceeds $\gamma R_t$, the algorithm suppresses the norm, while fidelity improvements are driven only by samples with small score norms. Let $\mathcal{S}\subset\mathcal{T}$ denote the subset of training data $\mathcal{T}$ satisfying $\frac{\beta_t}{2}\|\widehat{\bv}_t(\bx_{t,i})\|_2^2 > \gamma R_t$. With corresponding small $\lambda$, the standard diffusion loss $\mathcal{L}_{\epsilon}(\bx_{t,i},t)$, the training objective at time step $t$ is modified as 
\begin{equation}
\begin{aligned}
    \mathcal{L}_{\mathrm{train}}(t)
    =
    \frac{1}{N}
    \Bigg[
    \lambda
    \sum_{\bx_i\in\mathcal{S}}
    \left(
    \frac{\beta_t}{2}
    \left\|
    \widehat{\bv}_t(\bx_{t,i})
    \right\|_2^2
    -
    \gamma R_t
    \right)
    +
    \sum_{\bx_i\in\mathcal{T}\backslash\mathcal{S}}
    \mathcal{L}_{\epsilon}(\bx_{t,i},t)
    \Bigg].
\end{aligned}
\end{equation}
Here, the term $\lambda(\frac{\beta_t}{2}\left\|\widehat{\bv}_t(\bx_{t,i})\right\|_2^2 - \gamma R_t)$ acts as a norm-suppressing penalty. Smaller values of $\gamma$ place greater emphasis on norm suppression.


\begin{table}[t]
    \centering
    \footnotesize
    \setlength{\tabcolsep}{4pt}
    \renewcommand{\arraystretch}{1.12}
    \caption{\textbf{CelebA-HQ fine-tuning with velocity norm suppression.}
    Norm-suppressing penalty replaced the fine-tuning objective once the norm exceeds a threshold. Suppression is applied to 
    $\widehat{\bv}_t=x_t+\bs_\theta(x_t,t)$ and to the score
    $\bs_\theta(x_t,t)$ alone.
    }
    \vspace{-0.45em}
    \label{tab:reference_scale_finetuning}
    \begin{tabular}{lccc@{\hspace{1.4em}}ccc}
        \toprule
        Condition & \multicolumn{2}{c}{LPIPS $\leq 0.1$} & \multicolumn{2}{c}{LPIPS $\leq 0.3$} & \multicolumn{2}{c}{FID} \\
        \midrule
        Pretrained model & \multicolumn{2}{c}{0.00\%} & \multicolumn{2}{c}{0.08\%} & \multicolumn{2}{c}{14.93} \\
        No reference scale & \multicolumn{2}{c}{28.53\%} & \multicolumn{2}{c}{33.11\%} & \multicolumn{2}{c}{24.33} \\
        \midrule
        &
        \multicolumn{3}{c}{\textbf{Full velocity }$\widehat{\bv}_t=x_t+\bs_\theta(x_t,t)$\ \ {|\ \tiny$\lambda=10^{-3}$} } &
        \multicolumn{3}{c}{\textbf{Score-only }$\widehat{\bv}_t=\bs_\theta(x_t,t)$\ \ {\tiny$|\ \lambda=10^{-5}$}} \\
        \cmidrule(lr){2-4}\cmidrule(lr){5-7}
        Condition
        & LPIPS $\leq 0.1$ & LPIPS $\leq 0.3$ & FID
        & LPIPS $\leq 0.1$ & LPIPS $\leq 0.3$ & FID \\
        \midrule
        $\gamma=1.0$
        & 31.85\% & 36.79\% & 26.19
        & 23.14\% & 31.59\% & 22.71 \\
        $\gamma=0.8$
        & 11.39\% & 28.14\% & 17.99
        & 2.47\% & 18.01\% & 53.80 \\
        $\gamma=0.7$
        & 3.50\% & 20.65\% & 15.88
        & 1.46\% & 15.75\% & 25.74 \\
        $\gamma=0.5$
        & 0.05\% & 15.31\% & \textbf{14.38}
        & 0.29\% & 2.31\% & 156.08 \\
        $\gamma=0.3$
        & \textbf{0.00\%} & \textbf{3.80\%} & 15.87
        & 0.02\% & 0.23\% & 295.58 \\
        \bottomrule
    \end{tabular}%
    \vspace{-1.5em}
\end{table}


As shown in Table~\ref{tab:reference_scale_finetuning}, norm suppression effectively reduces nearest-neighbor replication in the repeated-sample fine-tuning setting. While it mitigates memorization, it does not hinder the improvement of the quality of generated images. 
For the full-velocity suppression objective, with $\gamma=0.5$, the replication rate drops from $28.53\%$ to $0.05\%$, while the FID improves from $24.33$ to $14.38$. Further suppression with $\gamma=0.3$ continues to reduce replication to $0.00\%$, but at the cost of slightly degraded fidelity. Learning with score penalizing compromises fidelity and fails to improve both  LPIPS and FID.

\subsection{Large-Scale Memorization Mitigation}
\label{subsec:large_scale_mitigation}

We further evaluate the proposed regularization method
on the large-scale LAION-Aesthetics-12M-UMAP dataset
using Stable Diffusion v1.4.
We compare our method with the memorization mitigation
approach of Wen et al.~\cite{wen2024detecting}.
Memorization is evaluated using the fraction of
generated samples with SSCD similarity exceeding 0.7,
while generation fidelity is assessed using CLIP scores.

\begin{table}[b]
\centering
\footnotesize
\setlength{\tabcolsep}{6pt}
\renewcommand{\arraystretch}{1.12}
\caption{
\textbf{Memorization mitigation on LAION-Aesthetics-12M-UMAP.}
Comparison of regularization methods on Stable Diffusion
v1.4 using 200 generated samples.
$\Delta$CLIP denotes the change in CLIP score relative
to the unmitigated model.
}
\vspace{-0.45em}
\label{tab:large_scale_mitigation}
\begin{tabular}{lcccc}
\toprule
\textbf{Method}
& \textbf{SSCD $>0.7$}
& \textbf{Count}
& \textbf{CLIP}
& \textbf{$\Delta$CLIP} \\
\midrule
No mitigation
& 39.00\%
& 78
& 0.3019
& -- \\

Wen ($\tau=2$) \cite{wen2024detecting}
& \textbf{0.00\%}
& \textbf{0}
& 0.2917
& -0.0102 \\

Wen ($\tau=4$)
& 0.50\%
& 1
& 0.2978
& -0.0041 \\

\midrule
Ours ($\gamma=0.5,\lambda=10^{-3}$)
& \textbf{0.00\%}
& \textbf{0}
& \textbf{0.2987}
& \textbf{-0.0032} \\
\bottomrule
\end{tabular}
\end{table}

As shown in Table~\ref{tab:large_scale_mitigation},
the unmitigated model produces 78 memorized samples
out of 200 generations (39.00\%).
Our method reduces the observed memorization rate to
0.00\%, matching Wen et al.\ ($\tau=2$), while
achieving a higher CLIP score (0.2987 vs.\ 0.2917).
Furthermore, our method incurs a smaller reduction
in CLIP score (-0.0032) than either configuration
of Wen et al.
These results suggest that the proposed regularization
can effectively mitigate memorization while better
preserving generation fidelity in this evaluation.

\section{Conclusion}
\label{sec:conclusion}

We introduced a thermodynamic diagnostic for model-level memorization in diffusion networks. The proposed criterion is based on a thermodynamic quantity---the rate of change of the Helmholtz free energy---computed from the learned score. Across a range of settings exhibiting memorization and non-memorization, networks for which this criterion does not diverge do not produce memorized samples. The proposed criterion achieves competitive performance compared with existing methods and can further mitigate memorization when incorporated into the training loss.

\bibliographystyle{abbrvnat}   
\bibliography{Helmholtz}       

\appendix

\section*{Appendix}

\section{Experimental Details}
\label{app:experiment_details}

The details of the experiment are as follows: Figure~\ref{fig:gaussian_experiment}(A), \ref{fig:cifar_finetuning}, and \ref{fig:celeba_ldm}
provides the plot of $\Phi=\frac{1}{T}\frac{dF}{dt}$ at each $t$ calculated using $\bv_t(\bx_t) = \bx + \bs_\theta(\bx_t, t)$. The experiments have been performed with the network estimating the total accumulated error $\epsilon_{\mathrm{tot}, t} = \mathbb{E}_{\rho_t(\bx, \bx_t)}[\bx_t - \bx]$ with training sample $\bx$ and the forward diffused sample $\bx_t$ at time $t$. The neural network in our experiments estimates $\epsilon_\theta(\bx_t,t)$ as in \cite{ho2020denoising}, and the learned score is obtained using $\bs_\theta(x_t,t)=-\epsilon_\theta(\bx_t,t)/\sigma_t$ with  $\sigma_t=\sqrt{1-e^{-\beta t}}$ or $\sigma_t = \sqrt{1 - (1 - \beta\Delta t)^{t/\Delta t}}$ for discrete algorithm. In all experiments, the diffusion timestep axis is shown on a logarithmic scale to enhance the visibility of the low-noise region near $t=0$.

The reverse process begins at $t = 1{,}000$ in our experiments, and data are generated at $t = 0$. The low-noise region lies near $t = 0$, where noise has accumulated through the forward process.

\subsection{Two-Dimensional Gaussian Details}
\label{app:gaussian_details}

\textbf{Affine network learns scores that generate Gaussian data}: The derivation below shows that this network recovers the score corresponding to a Gaussian with maximum-likelihood estimates of the mean and covariance. 

We consider the following score matching objective function:
\begin{eqnarray}
    L(\bs) = \sum_{i=1}^N \frac{1}{2}\bs(\bx_i)^2 + \nabla\cdot\bs(\bx_i),
\end{eqnarray}
which is a modified expression using the integration by parts from the mean squared error between the estimated score and the true score \cite{hyvarinen2005score,song2021scorebased}. Consider an affine score function $\bs(\bx, t)=A(\bx - \bm)$ where $\bs(\bx, t)\in\RE^d$, $\bx\in\RE^d$, and $t\in\RE$. Plugging $\bs(\bx) = A(\bx - \bm)$ into $L$ yields $L=\sum_i\frac{1}{2}(\bx-\bm)^\top AA(\bx-\bm) + tr[A]$. The optimal $A$ and $\bm$ that minimize $L$, obtained from the conditions $\frac{\partial L}{\partial A} = 0$ and $\frac{\partial L}{\partial \bm} = 0$, are given in closed form as
\begin{eqnarray}
    A_* = -\left[\frac{1}{N}\sum_{i=1}^N(\bx_i - \bm)(\bx_i - \bm)^\top\right]^{-1}\!\!\!\equiv -\widehat{\Sigma}^{-1}, \quad 
    \bm_* = \frac{1}{N}\sum_{i=1}^N \bx_i \equiv \widehat{\mu},
\end{eqnarray}
where $\widehat{\mu}$ and $\widehat{\Sigma}$ denote the empirical mean and covariance, respectively. The resulting score corresponds to that of a Gaussian with mean $\widehat{\mu}$ and covariance $\widehat{\Sigma}$, since the Gaussian score is $\nabla\log\rho = -\Sigma^{-1}(\bx - \mu)$ for $\rho(\bx) = \mathcal{N}(\bx; \mu, \Sigma)$. Therefore, Langevin dynamics using an affine score model generates Gaussian data with the empirical mean and covariance of the training data, regardless of whether the original data are Gaussian. \hfill $\blacksquare$

The experiment was conducted on a two-dimensional Gaussian distribution with zero mean. The covariance matrix is constructed by rotating a diagonal matrix with entries $(4.0, 0.25)$ by $35^\circ$. We use $N = 10$ samples from this distribution for training. Subsequently, the Kullback–Leibler (K-L) divergence is computed using the generated samples and the analytical form of the underlying density.

We use $1{,}000$ time steps for both the forward and reverse processes, with a constant $\beta$ of $0.0095$. 
Each model is trained for $20{,}000$ optimization steps using AdamW, an Adam variant with decoupled weight decay \cite{loshchilov2019decoupled}, with the standard $\epsilon$-prediction loss, a batch size of $256$, and a learning rate of $10^{-3}$.
Figure~\ref{fig:gaussian_experiment} presents a comparison of model capacity for $N = 10$ between an affine denoiser, a small MLP, and a large MLP.

The affine denoiser uses a 3-layer time MLP with 64 hidden nodes and has 10,757 trainable parameters.
The small MLP has 2 hidden layers with 28 hidden nodes, resulting in 1,850 trainable parameters, whereas the large MLP has 6 hidden layers with 1,024 hidden nodes and 5,285,890 trainable parameters.
Thus, the small MLP has limited capacity, while the large MLP has sufficient capacity to form sample-specific representations around the finite training set.

The small and large MLP denoisers share the same input variables but differ in capacity:
\begin{equation}
    \epsilon_\theta^{\mathrm{MLP}}(\bx_t,t)
    =
    \mathrm{MLP}_\theta
    \left(
    [\bx_t,\gamma(t)]
    \right),
    \label{eq:mlp_gaussian_denoiser}
\end{equation}
where $\gamma(t)$ is a sinusoidal time embedding.


The affine denoiser has the following formulation:
\begin{equation}
    \epsilon_\theta^{\mathrm{affine}}(\bx_t,t)
    =
    A_\theta(t)
    \left(
    \bx_t-m_\theta(t)
    \right),
    \label{eq:affine_gaussian_denoiser}
\end{equation}
where $A_\theta(t)$ is a symmetric matrix and both $A_\theta(t)$ and $m_\theta(t)$ are produced from the time embedding. Score has been obtained using $s_\theta(\bx_t,t)=-\epsilon^{\mathrm{affine}}_\theta(\bx_t,t)/\sigma_t$, Eq.~(\ref{eq:affine_gaussian_denoiser}). 

In Figure~\ref{fig:gaussian_experiment}(B), we compare the generated distribution of each model with the true underlying Gaussian density.
For each trained model, we first generate samples from the model distribution $p_\theta$ and fit a Gaussian KDE, denoted by $\hat{p}^{\mathrm{KDE}}_\theta$, to these samples.
We then draw evaluation samples $\left\{\bx_i^{\mathrm{eval}}\right\}_{i=1}^{N}$ from the fitted KDE and compute the following Monte Carlo estimate of the divergence to the true density:
\begin{equation}
    \widehat{\mathrm{KL}}
    \left(
    \hat p_\theta^{\mathrm{KDE}}
    \,\|\,
    p_{\mathrm{true}}
    \right)
    =
    \frac{1}{N}
    \sum_{i=1}^{N}
    \left[
    \log \hat{p}^{\mathrm{KDE}}_\theta
    \left(
    \bx_i^{\mathrm{eval}}
    \right)
    -
    \log p_{\mathrm{true}}
    \left(
    \bx_i^{\mathrm{eval}}
    \right)
    \right],
    \qquad
    \bx_i^{\mathrm{eval}} \sim \hat p_\theta^{\mathrm{KDE}} .
\end{equation}

\subsection{Real-World Experiment Using Benchmark Dataset}

For CIFAR-10 and CelebA-HQ, fine-tuning updates the pretrained model for $50{,}000$ steps using a subset of the data. This subset is repeatedly presented to the model to promote memorization. Under the subset-to-full update ratio setting, $5{,}000$ randomly sampled images from the full dataset are intermittently introduced every $\tau$ steps. The CIFAR-10 experiment measures statistics in RGB pixel space, whereas the CelebA-HQ experiment uses the latent denoising space provided by the pretrained model. The resolution of CIFAR-10 is $32 \times 32$, while that of CelebA-HQ is $256 \times 256$. After fine-tuning, nearest-neighbor LPIPS and FID are computed using the full training dataset.

\section{Samples of Generated Images}
\label{app:qualitative_samples}

\begin{figure}[t]
    \centering
    \begin{minipage}{0.45\textwidth}
        \centering
        \textbf{Memorized: $N=100$}\par
        \includegraphics[width=\linewidth,keepaspectratio]{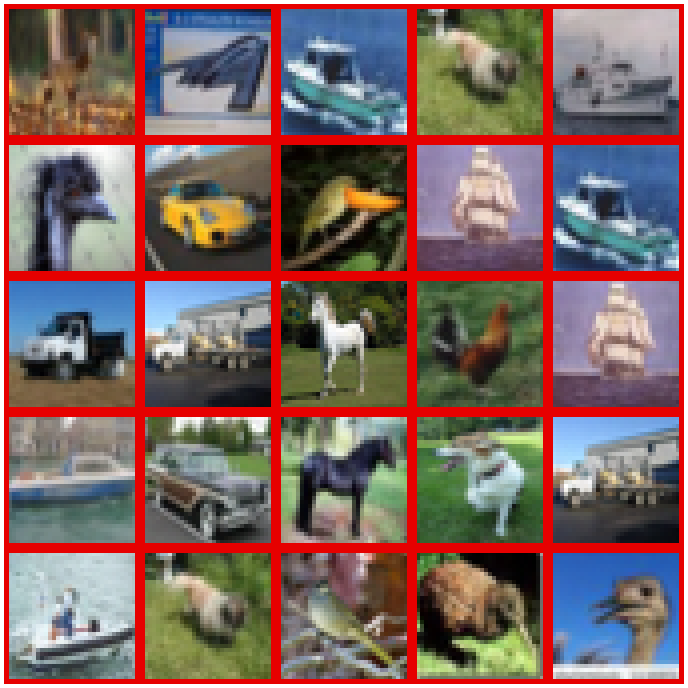}
    \end{minipage}
    \hfill
    \begin{minipage}{0.45\textwidth}
        \centering
        \textbf{Less Memorized: $N=20{,}000$}\par
        \includegraphics[width=\linewidth,keepaspectratio]{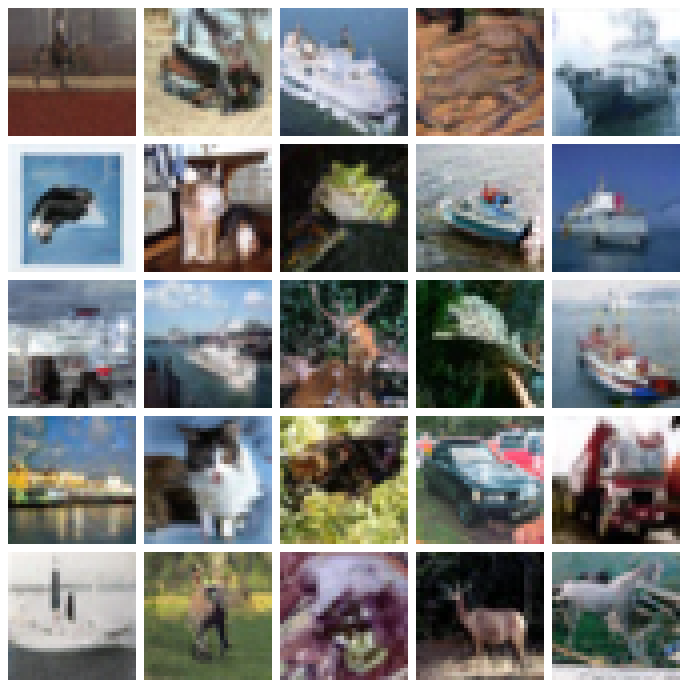}
    \end{minipage}

    \begin{minipage}{0.45\textwidth}
        \centering
        \textbf{Memorized: $\tau=100$}\par
        \includegraphics[width=\linewidth,keepaspectratio]{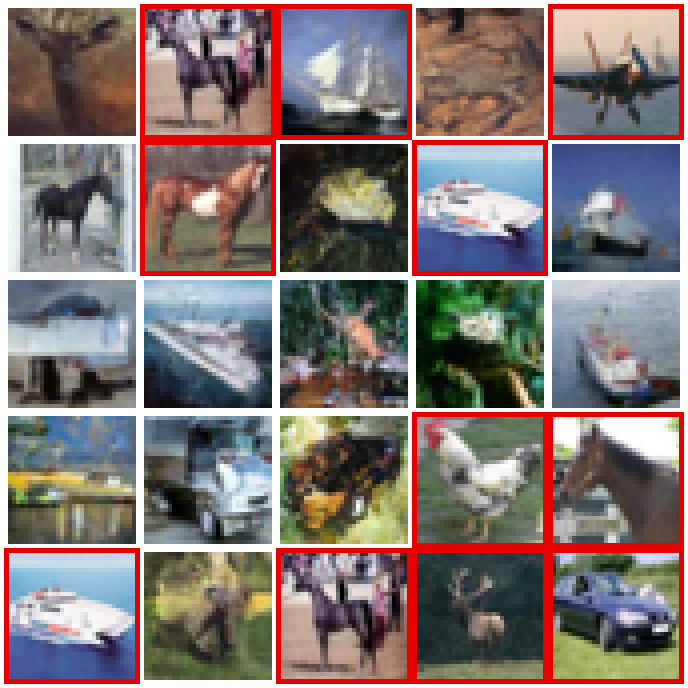}
    \end{minipage}
    \hfill
    \begin{minipage}{0.45\textwidth}
        \centering
        \textbf{Less Memorized: $\tau=0$}\par
        \includegraphics[width=\linewidth,keepaspectratio]{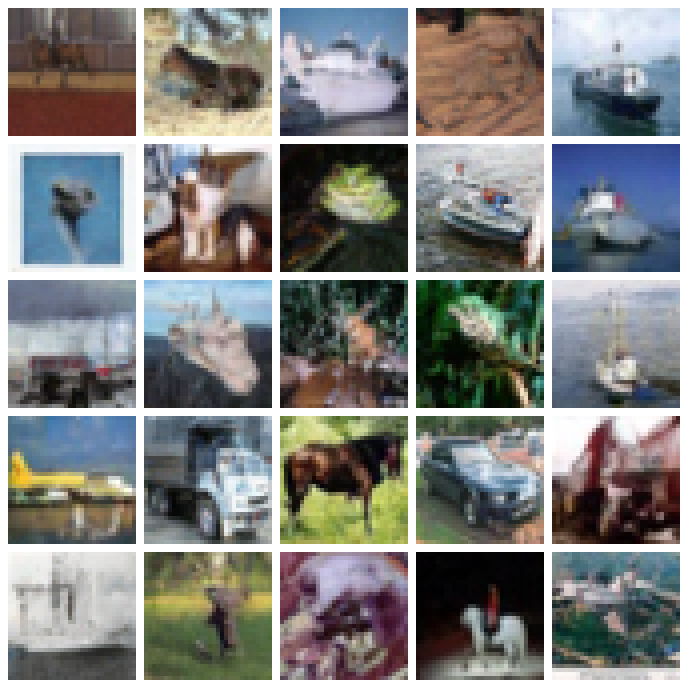}
    \end{minipage}
    \caption{
    \textbf{CIFAR-10 generated examples.}
    Top: catastrophic corruption conditions with learned subset size $N_s$.
    Bottom: non-catastrophic corruption conditions with subset-to-full update ratio $\tau$.
    Red boxes indicate that the generated samples are flagged as nearest-neighbor replication  under the criterion LPIPS $\le 0.03$.
    }
    
    \label{fig:app_cifar_samples}
\end{figure}

\begin{figure}[t]
    \centering
    \begin{minipage}{0.45\textwidth}
        \centering
        \textbf{Memorized: $N=100$}\par
        \includegraphics[width=\linewidth,keepaspectratio]{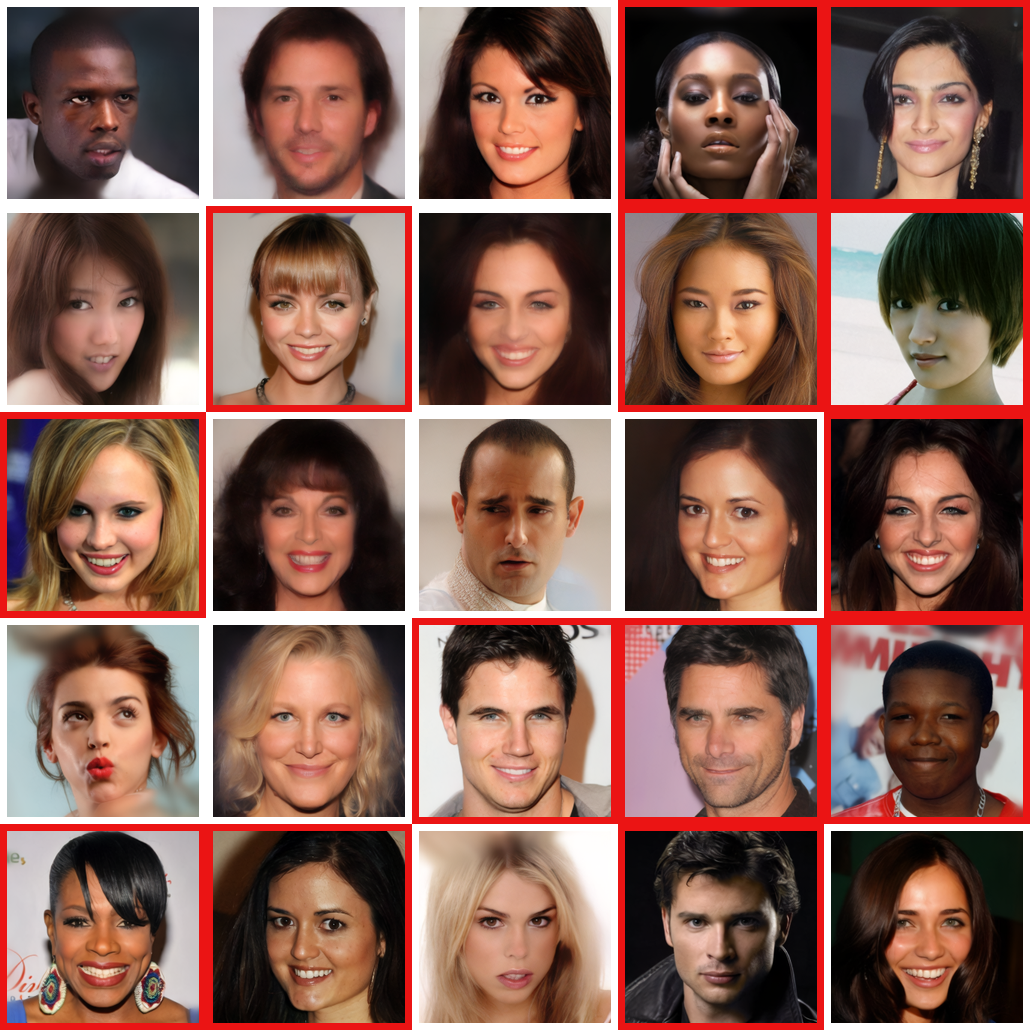}
    \end{minipage}
    \hfill
    \begin{minipage}{0.45\textwidth}
        \centering
        \textbf{Less Memorized: $N=20{,}000$}\par
        \includegraphics[width=\linewidth,keepaspectratio]{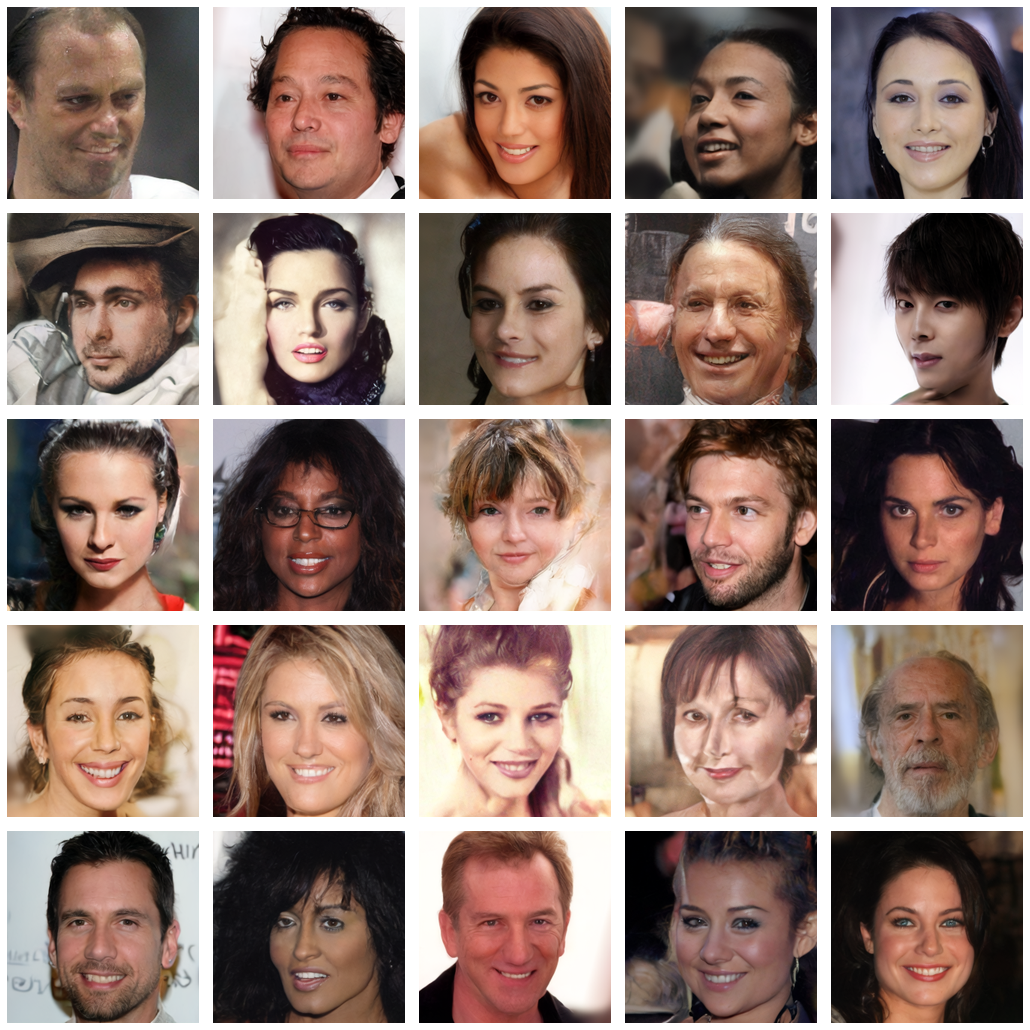}
    \end{minipage}

    \begin{minipage}{0.45\textwidth}
        \centering
        \textbf{Memorized: $\tau=100$}\par
        \includegraphics[width=\linewidth,keepaspectratio]{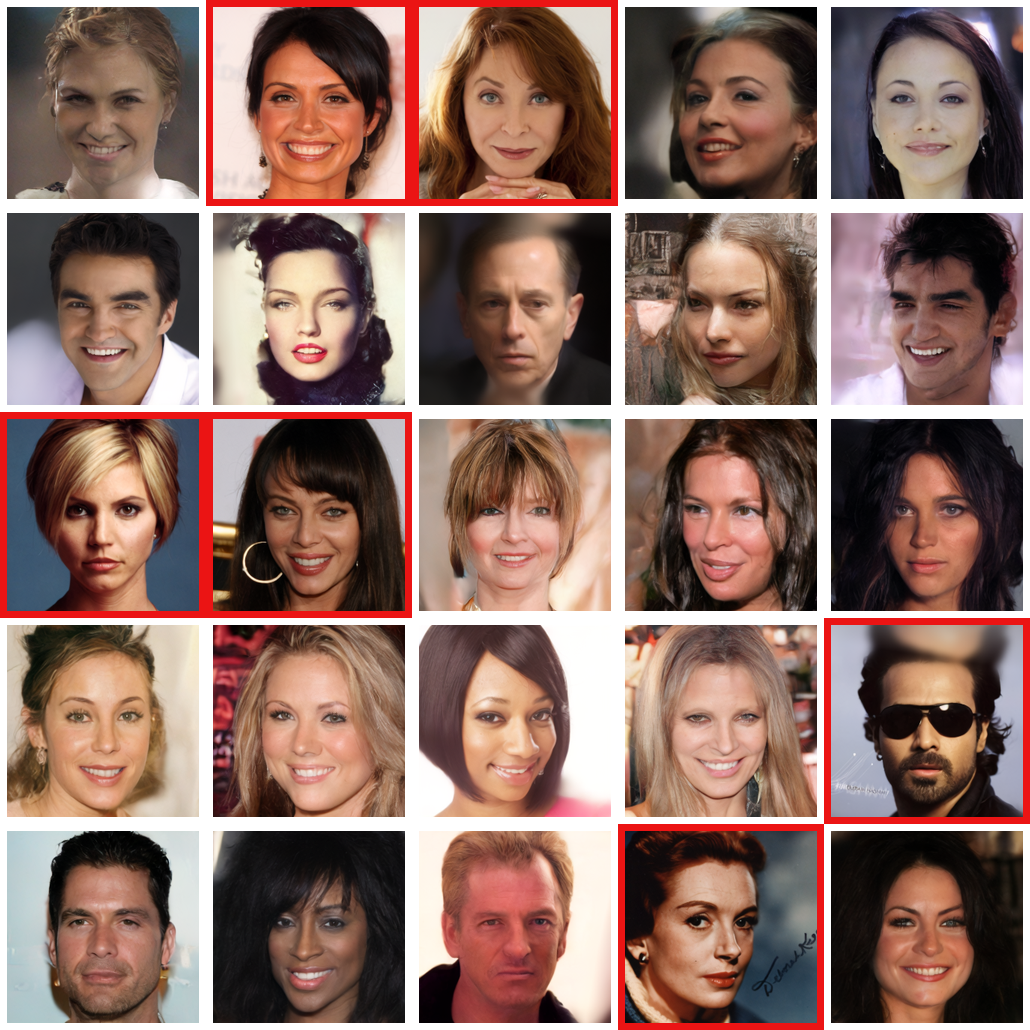}
    \end{minipage}
    \hfill
    \begin{minipage}{0.45\textwidth}
        \centering
        \textbf{Less Memorized: $\tau=0$}\par
        \includegraphics[width=\linewidth,keepaspectratio]{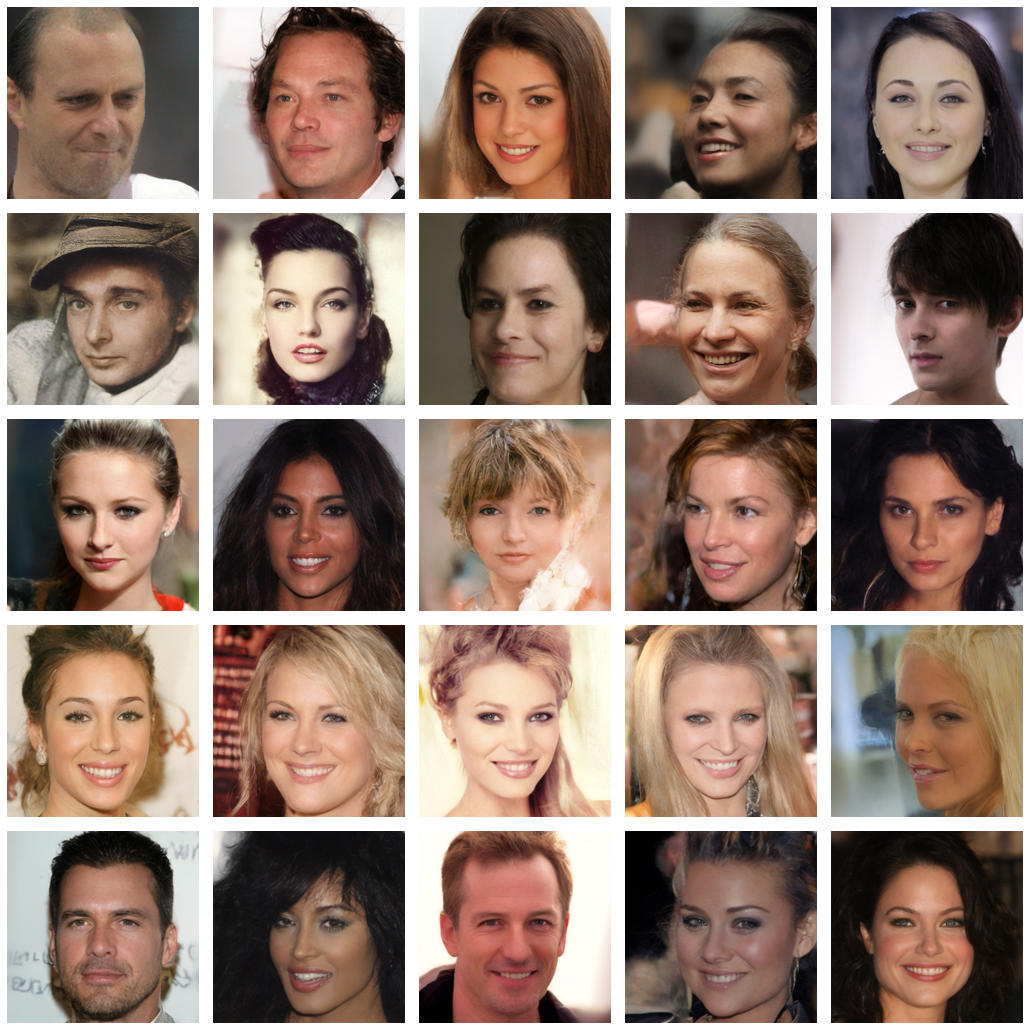}
    \end{minipage}
    \caption{
    \textbf{CelebA-HQ generated examples.}
    Top: catastrophic corruption conditions with learned subset size $N_s$.
    Bottom: non-catastrophic corruption conditions with subset-to-full update ratio $\tau$.
    Red boxes indicate that the generated samples are flagged as nearest-neighbor replication  under the criterion LPIPS $\le 0.1$.
    }
    
    \label{fig:app_celeba_samples}
\end{figure}

We present examples of generated samples from the real-world experiments. Settings that promote memorization are shown on the left, while settings that discourage memorization are shown on the right, for both the CIFAR-10 in Figure~\ref{fig:app_cifar_samples} and CelebA-HQ in Figure~\ref{fig:app_celeba_samples}. Red boxes indicate images with LPIPS scores below the memorization threshold used in Table~\ref{tab:lpips_memorization}: LPIPS $\le 0.03$ for CIFAR-10 and LPIPS $\le 0.1$ for CelebA-HQ.

In $N_s=100$ experiment, all samples in CIFAR-10 and many samples in CelebA-HQ are highlighted with red boxes, indicating severe memorization by the networks. Under the non-catastrophic corruption setting with $\tau = 100$, many examples are also red-boxed, although they do not constitute the majority. In contrast, for the non-memorizing settings ($N_s = 20{,}000$ and $\tau = 0$), none of the generated samples exhibit memorization. These observations are consistent with the norm of the rate of Helmholtz free energy dissipation shown in Figures~\ref{fig:cifar_finetuning} and~\ref{fig:celeba_ldm}.

\FloatBarrier

\FloatBarrier

\section{Sample-Level Memorization in Stable Diffusion 2 Base}
\label{app:sample-level_memorization_in_SD2_base}

We present sample-level memorization detection results on Stable Diffusion 2 Base. The same 1,000-step DDPM measurement protocol used in Section~\ref{subsec:sample_level_detection} is applied. As in the Stable Diffusion v1.4 experiments, memorization labels are assigned based on prompts categorized as memorization-prone or non-memorization-prone, rather than verified replication of training images.

\begin{table}[!htbp]
\centering
\footnotesize
\setlength{\tabcolsep}{5pt}
\renewcommand{\arraystretch}{1.12}
\caption{
\textbf{Sample-level memorization detection on Stable Diffusion 2 Base.}
All measurements are performed on DDPM 1,000-step trajectories.
Bold entries are statistically indistinguishable from the best value in each column under an approximate two-sided test at $p=0.05$ using the reported standard errors.
}
\label{tab:sample_level_auroc_sd20}
\begin{tabular}{lccc}
\toprule
\textbf{Method}
& \textbf{Time step}
& \textbf{AUC}
& \textbf{TPR@1\%FPR} \\
\midrule

$\left\|\bs_\theta^\Delta(\bx_t)\right\|$
& 1,000
& \separen{0.977}{0.002}
& \separen{0.822}{0.010} \\

$= \|\bs_{\theta,\mathrm{cond}}-\bs_{\theta,\mathrm{uncond}}\|$ \cite{wen2024detecting}
& 1
& \bseparen{0.996}{0.000}
& \bseparen{0.967}{0.002} \\

\midrule

$\left\|H_\theta^\Delta(\bx_T)\bs_\theta^\Delta(\bx_T)\right\|^2$ \cite{jeon2025understanding}
& 1,000
& \separen{0.969}{0.002}
& \separen{0.562}{0.028} \\

& 1
& \separen{0.897}{0.002}
& \separen{0.001}{0.001} \\

\midrule

$\|\widehat{\bv}_{\mathrm{cfg}}\|_2^2-\|\widehat{\bv}_{\mathrm{uncond}}\|_2^2$ (Ours)
& 1
& \separen{0.988}{0.001}
& \separen{0.927}{0.006} \\

\bottomrule
\end{tabular}
\end{table}

\FloatBarrier

\section{Derivation of the Threshold for the Loss Function Modification}
\label{sec:Threshold}

In this section, we derive the threshold $R_t$ used in Section~\ref{sec:mitigating_memorization_reference_scale}. The threshold is chosen as the value of the penalized quantity $\frac{\beta_t}{2}\|\widehat{\bv}_t\|_2^2$ when the network has fully memorized the training data, i.e., when the learned score coincides with the score of a single diffused $\delta$-function.

\textbf{Forward process and learned velocity:} Accumulating the forward update in Eq.~(\ref{eq:forward_pr}) over $t$ steps, a training sample $\bx_i\in\RE^d$ is diffused to
\begin{eqnarray}
    \bx_{t,i} = \sqrt{1-\sigma_t^2}\,\bx_i + \sigma_t\epsilon, \quad \epsilon\sim\mathcal{N}(\epsilon; 0, I),
    \label{eq:forward_marginal}
\end{eqnarray}
where $\sigma_t^2 = 1-\prod_{s=0}^{t-1}(1-\beta_s)$ is the accumulated variance. The network predicts $\epsilon_\theta(\bx_t,t)$, from which the learned score is $\bs_\theta(\bx_t,t) = -\epsilon_\theta(\bx_t,t)/\sigma_t$. Following Eq.~(\ref{eq:velocity_forward}), the velocity field is $\bv = -\frac{\beta_t}{2}(\bx + \nabla\log\rho_t)$, and its learned counterpart without the common factor $-\frac{\beta_t}{2}$ is
\begin{eqnarray}
    \widehat{\bv}_t(\bx_t) = \bx_t + \bs_\theta(\bx_t,t) = \bx_t - \frac{\epsilon_\theta(\bx_t,t)}{\sigma_t}.
\end{eqnarray}

\textbf{Point-mass score:} As shown in Section~\ref{sec:delta_function_diffusion}, the velocity field around a training sample is dominated by the nearest $\delta$-function. A memorizing network therefore reproduces, near $\bx_{t,i}$, the score of the diffused $\delta$-function centered at $\bx_i$, which is the Gaussian $\mathcal{N}(\bx; \sqrt{1-\sigma_t^2}\,\bx_i, \sigma_t^2 I)$:
\begin{eqnarray}
    \nabla\log\rho_t(\bx_{t,i}) = -\frac{\bx_{t,i}-\sqrt{1-\sigma_t^2}\,\bx_i}{\sigma_t^2} = -\frac{\epsilon}{\sigma_t},
\end{eqnarray}
where the last equality follows from Eq.~(\ref{eq:forward_marginal}). This is exactly the score obtained when the network predicts the injected noise without error, $\epsilon_\theta(\bx_{t,i},t)=\epsilon$.

\textbf{Point-mass velocity:} Substituting the point-mass score and Eq.~(\ref{eq:forward_marginal}) into $\bx_{t,i} + \nabla\log\rho_t(\bx_{t,i})$ gives
\begin{eqnarray}
    \bx_{t,i} + \nabla\log\rho_t(\bx_{t,i})
    &=& \sqrt{1-\sigma_t^2}\,\bx_i + \left(\sigma_t - \frac{1}{\sigma_t}\right)\epsilon \nonumber\\
    &=& \sqrt{1-\sigma_t^2}\,\bx_i - \frac{1-\sigma_t^2}{\sigma_t}\,\epsilon.
\end{eqnarray}
Taking the expectation over $\epsilon$, the cross term vanishes since $\mathbb{E}[\epsilon]=0$, and $\mathbb{E}\|\epsilon\|_2^2=d$. The expected penalized quantity for the sample $\bx_i$ is then
\begin{eqnarray}
    \frac{\beta_t}{2}\,\mathbb{E}_{\epsilon}\!\left[\left\|\bx_{t,i} + \nabla\log\rho_t(\bx_{t,i})\right\|_2^2\right]
    &=& \frac{\beta_t}{2}\left( (1-\sigma_t^2)\|\bx_i\|_2^2 + \frac{(1-\sigma_t^2)^2}{\sigma_t^2}\,d \right) \nonumber\\
    &=& \frac{\beta_t}{2}(1-\sigma_t^2)\left(\|\bx_i\|_2^2 + \frac{1-\sigma_t^2}{\sigma_t^2}\,d\right).
    \label{eq:pointmass_reference}
\end{eqnarray}

\textbf{Threshold:} Eq.~(\ref{eq:pointmass_reference}) depends on the individual sample only through $\|\bx_i\|_2^2$. For normalized data with unit variance per dimension, $\|\bx_i\|_2^2\approx d$; this holds for the scaled latents of the latent diffusion model. Under this approximation, the two terms combine as $1 + \frac{1-\sigma_t^2}{\sigma_t^2} = \frac{1}{\sigma_t^2}$, and the sample-independent threshold becomes
\begin{eqnarray}
    \boxed{
    R_t = \frac{\beta_t(1-\sigma_t^2)}{2\sigma_t^2}\,d.
    }
\end{eqnarray}
For data that are not normalized, $\|\bx_i\|_2^2$ in Eq.~(\ref{eq:pointmass_reference}) can be replaced by its dataset average $\frac{1}{N}\sum_{i=1}^N\|\bx_i\|_2^2$.

A training sample $\bx_i$ is assigned to the subset $\mathcal{S}$ in Section~\ref{sec:mitigating_memorization_reference_scale} when
\begin{eqnarray}
    \frac{\beta_t}{2}\left\|\widehat{\bv}_t(\bx_{t,i})\right\|_2^2 > \gamma R_t,
\end{eqnarray}
i.e., when the learned velocity norm exceeds the fraction $\gamma$ of the point-mass reference. Since the penalty acts only on such samples, it prevents the network from approaching the singular velocity of the memorized reverse process, while the remaining samples are trained with the standard loss.

\end{document}